\documentclass[runningheads]{llncs}

\usepackage{graphics}
\usepackage{soul}
\usepackage{tcolorbox}
\usepackage{glossaries}
\usepackage{soul}
\usepackage{amsmath}
\usepackage{bbm}
\usepackage{amsfonts}
\usepackage{siunitx}

\usepackage{caption}
\newacronym{rl}{RL}{Reinforcement Learning}
\newacronym{smdp}{SMDP}{Semi-Markov Decision Process}
\newacronym{mdp}{MDP}{Markov Decision Process}
\newacronym{pomdp}{POMDP}{Partially Observable Markov Decision Process}
\newacronym{hrl}{HRL}{Hierarchical Reinforcement Learning}
\newacronym{ppo}{PPO}{Proximal Policy Optimization}
\newacronym{gru}{GRU}{Gated Recurrent Unit}
\newacronym{setie}{SETIE}{Structured Exploration Through Instruction Enhancement}

\begin{document}
\title{Structured Exploration Through Instruction Enhancement for Object Navigation}
\titlerunning{Structured Exploration Through Instruction Enhancement}

\author{Matthias Hutsebaut-Buysse \orcidID{0000-0001-6091-294X}\and  \\
Tom De Schepper \orcidID{0000-0002-2969-3133} \and \\
Kevin Mets \orcidID{0000-0002-4812-4841} \and \\
Steven Latr\'e \orcidID{0000-0003-0351-1714}
}

\authorrunning{M. Hutsebaut-Buysse et al.}

\institute{University of Antwerp - imec\\ IDLab - Department of Computer Science  \\
\email{\{firstname.lastname\}@uantwerpen.be}}

\maketitle              
\begin{abstract}
Finding an object of a specific class in an unseen environment remains an unsolved navigation problem. Hence, we propose a hierarchical learning-based method for object navigation. The top-level is capable of high-level planning, and building a memory on a \textit{floorplan}-level (e.g., which room makes the most sense for the agent to visit next, where has the agent already been?). While the lower-level is tasked with efficiently navigating between rooms and looking for objects in them. Instructions can be provided to the agent using a simple synthetic language. The top-level intelligently enhances the instructions in order to make the overall task more tractable. Language grounding, mapping instructions to visual observations, is performed by utilizing an additional separate supervised trained goal assessment module. We demonstrate the effectiveness of our method on a dynamic configurable domestic environment.

\end{abstract}
\section{Introduction}

Finding objects in unseen environments is a hard navigation task. In order to be successful, an agent needs to be capable of mastering a number of skills. First, the agent needs to be capable to explore the environment in a structured manner: it should figure out the layout of the previously unseen environment, keep a memory of past actions, and remember visited regions. Second, the agent needs to be capable to understand the instruction: map an instruction to an actual visual representation. Third, the agent needs to be capable to make decisions on multiple abstraction levels: navigate to the other side of the building versus navigating through a doorway. 

These problems have been studied individually intensively in various settings \cite{szot2021habitat2,weihs2020allenact,savva2019habitat,chevalierboisvert2019babyai}. However, constructing an agent capable of simultaneously performing these feats, remains an open challenge. In this paper we study how we can build an agent capable of simultaneously handling long-term planning through abstraction, low-level locomotion and basic language grounding.

Current navigation solutions typically utilize a \textit{sense-plan-act} approach, in which different modules interact with each other. These solutions however tend to be brittle, are prone to error propagation, and often require a lot of manual engineering \cite{karkus2021diff_slam,mishkin2019nav_benchmark}. End-to-end \gls{rl} systems have recently been proposed, as an alternative learning through interactions based solution, to handle these issues \cite{wijmans2020DDPPO}. Unfortunately, as we will demonstrate, \gls{rl} agents are often unable to reason on multiple levels of abstraction, have difficulties with mapping language instructions, and often explore poorly.

In contrast, our approach allows the agent to plan and explore on multiple levels of abstraction (e.g., on room-level and actuator-level) through utilizing a hierarchical approach. The proposed agent can be trained using only the reward-signal received from the environment, and only requires an egocentric RGB observation. This is in contrast to prior approaches, which often also require the pose of the agent as input. 

In order to communicate between the two layers we propose \textit{instruction enhancements}. In this system, the top-level is allowed to enhance the instruction it received from the environment. For example, if the original instruction is: \textit{"Find the red ball"}, the top-level might choose to enhance this instruction to: \textit{"Find the red ball, \underline{in the kitchen}"}. This allows the top-level to plan on a higher level of abstraction (Which room makes sense to visit next? Where have I already been?). In turn, the enhanced instruction makes the task more tractable to complete by the lower-level.

Because both traditional and learning-based approaches are still unsolved, we take one step back from the typically used photo-realistic simulators  \cite{szot2021habitat2,savva2019habitat}, and utilize a visually simpler setting \cite{gym_miniworld}, while keeping most of the navigation and generalization complexities. In this setting we demonstrate why a flat, non-hierarchical \gls{rl} agent, does not manage to make any progress, and how our hierarchical approach is capable of exploring the environment in a more principled way. We also demonstrate the generalization capabilities of the agent to find previously unseen objects in new unseen environment configurations.

The contributions of this work are three-fold: (1) We introduce a dual layer hierarchical approach, capable of simultaneously learning structured room-level exploration, and low-level navigation. (2) In order to communicate between layers we propose to enhance instructions, allowing loose coupling of layers and generalization to novel instructions. (3) The introduction of a goal assessment module, which is capable of addressing whether the current state satisfies the instruction, and thus allows offloading language grounding, and integration of prior knowledge in a learning-based setup.

\vspace{60pt}

\section{Background}

\textbf{Reinforcement Learning (RL)}\\
A sequential decision-making problem can be modelled as a \gls{pomdp}, represented by a tuple $\langle \mathcal{S},\mathcal{A},\mathcal{P},\mathcal{R}, \Omega, \mathcal{O}, \gamma \rangle$.

On each time step $t$, the agent samples an action $a_t \in \mathcal{A}$ from its policy $\pi(a_t|o_t, g_t)$, and the environment produces in turn an observation $o_t \in \Omega ,o_t \sim \mathcal{O}(s_t) $ of the internal state $s_t \in \mathcal{S}$ according to an unknown transition function $\mathcal{P}(s_{t+1}|s_t,a_t)$. The agent has access to a reward signal $r_t(s_t,a_t,g_t)$, which can be utilized to learn the value of the sequence of previously taken actions. In the goal-conditional \gls{rl} setting studied in this paper, the reward-signal depends on an additional goal-signal $g_t$ (the instruction). This goal-signal remains constant during each task instance (an episode). Episodes are terminated after a pre-determined step-limit is reached, or the agent utilizes a special \textit{done}-action.

The goal of \gls{rl} consists of finding a policy $\pi$ capable of maximizing the sum of rewards $R_t$, discounted by a factor $\gamma \in [0,1]$, through environment interactions: 

\begin{equation}
R_t = \underset{\pi, \mathcal{P}}{\mathbb{E}}\left[\sum_{t=0}^{T} \gamma^{t} r_{t}\left({o}_{t}, g_{t}, a_{t}, o_{t+1}\right)\right]
\end{equation} \\

\noindent \textbf{Proximal Policy Optimization (PPO)}\\
In order to learn a policy the on-policy \gls{ppo} algorithm \cite{schulman2017ppo} can be utilized. \gls{ppo} utilizes an importance-weighted advantage on samples collected in the environment during a rollout phase. A proximity clipping term is used as a trust region optimization method in order to allow updates to use experiences collected during a rollout multiple times. This is done in order to improve sample efficiency.
\\

\noindent \textbf{Hierarchical Reinforcement Learning (HRL)} \\
Exploration within policy-gradient methods such as \gls{ppo} is achieved through sampling actions from a stochastic policy. However, solely depending on this mechanism to find solutions for complex tasks is often not tractable \cite{nachum2019hrl_works}. 

Within a two-level goal-conditioned hierarchical approach, a meta-controller $\pi_{m}(z_{t}|o_{t},g_{t})$ maximizes the extrinsic reward signal $r_t$ indirectly by generating high-level actions $z_t \in \mathcal{Z}$ (often called skills or sub-behaviors). These high-level actions are executed for $c$ steps by a second low-level policy $\pi_{c}(a_t|o_{t}, z_{t})$, often called a controller. The controller maximizes an intrinsic reward signal by directly outputting primitive actions $a_t \in \mathcal{A}$.

\section{Approach}

\begin{figure}[t!]
    \centering
    \includegraphics[scale=0.8]{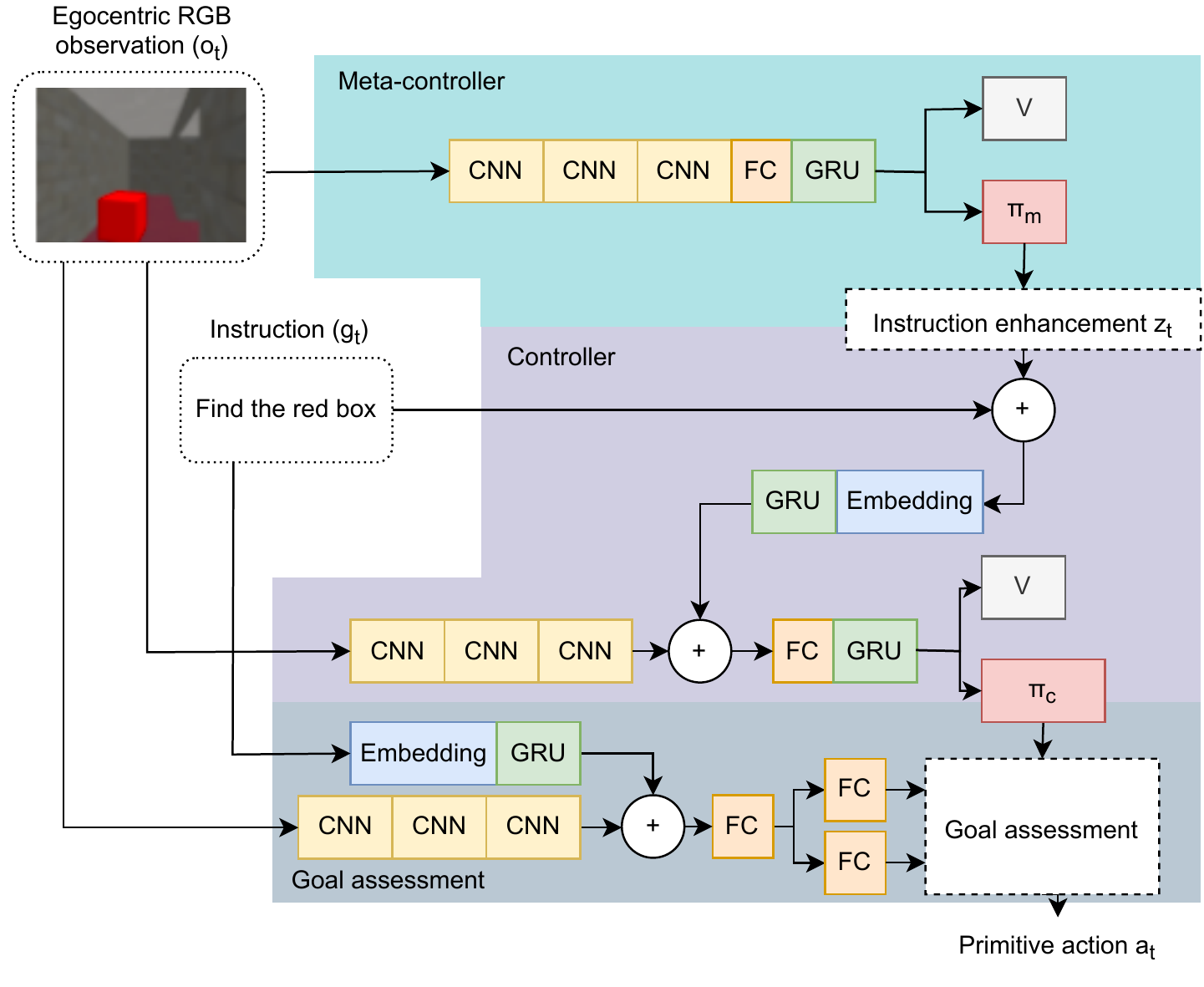}
    \caption{SETIE architecture: the meta-controller handles structured exploration between different rooms from egocentric observations by enhancing the instruction. This output is used by the controller, in order to return primitive actions (navigation). The goal assessment module is used for language grounding.}
    \label{fig:architecture}
\end{figure}

The proposed \gls{setie} approach consists of three parts: (a) the meta-controller $\pi_{m}(z_t|o_t,g_t)$ which performs high-level planning, by working on a lower temporal resolution, (b) the controller $\pi_{c}(a_t|o_t,g_t,z_t)$ which handles low-level navigation, and (c) the goal assessment module $G(o_t,g_t) \rightarrow \{1,0\}$ which handles language grounding. A visual representation of the architecture is displayed in Figure~\ref{fig:architecture}.

\subsection{Meta-controller}
The meta-controller $\pi_{m}(z_t|o_t,g_t)$ is responsible for learning high-level navigation of the environment solely from partial state observations (through an egocentric RGB camera). This task consists of two sub-tasks: (1) discovering the layout of the current environment, determining which rooms are connected to which other rooms. Commonsense reasoning (the garage is less likely to be connected with the bathroom) together with a trial-and-error approach can be used in order to solve this task. (2) Keeping an implicit memory of which rooms have already been visited in order to explore the environment in a structured manner. Because the meta-controller reasons on a higher level of abstraction, the agent is capable to perform these tasks using a generic \gls{gru} component \cite{cho2014gru} in its architecture.

The action-space of the meta-controller consists of a discrete set of \textit{instruction enhancements}. This set of enhancements is provided up-front to the agent. Instruction enhancements should be defined on a higher level of abstraction, than the primitive actions utilized by the controller. By introducing this additional level of abstraction, the agent is able to explore in a structured manner (e.g., room by room).

The meta-controller does not interact with the environment itself, but can only influence the behavior of the controller through enhancing the instruction. For example the extrinsic instruction $g_t$ could have been \textit{"Find the green key"}, which the meta-controller can enhance to become \textit{"Find the green key, in the dining room"}.

Within HRL, designing a sub-behavior space $\mathcal{Z}$ is a complex challenge. Most often this space is tightly coupled between the different levels. Utilizing language allows to decouple multiple levels. This allows the controller and meta-controller to be trained independently. Furthermore, language has also the potential to  generalize to unseen instructions \cite{jiang2019hal}, and can make the intention of the agent clear to a human in the loop \cite{chenAskYourHumans2021}.

The meta-controller acts on a lower temporal resolution and is asked to provide a new instruction enhancement every $c$ timesteps.

As the meta-controller has no direct influence on the environment, but only can act through the controller, its training needs to take into account potential unexpected behavior of a trained controller. Such quirks might be over-exploration of some rooms, while quickly moving through others. Accounting for these eccentricities can be done by utilizing a fully trained and frozen controller during training of the meta-controller. In this setting the meta-controller observes the environment, selects an instruction enhancement, and waits until the controller has taken $c$-steps, before sampling a novel enhancement. The reward of the meta-controller consists of the discounted sum of the extrinsic reward collected during the usage of the active instruction enhancement:
\begin{equation}
R_t(s_t)=1/c\sum_{t=0}^{c} \gamma^{t} r_{t}\left({o}_{t}, g_{t}, a_{t}, o_{t+1}\right)
\end{equation}
A second option to train the meta-controller consists of assuming a perfectly behaving controller. In this setting the (simulated) environment will carry out the enhancements, and move the agent to different rooms, while respecting the floor plan. Utilizing this second approach allows both controller and meta-controller to be trained in parallel (as there is no dependency). In order to utilize this second training scheme a different reward function is required. For example, a reward function based on the room coverage can be utilized. In this setting each instruction enhancement which takes the agent to a previously unvisited room will lead to a positive reward (0.1), while other proposed enhancements will result in a negative slack penalty (-0.01).

While in the empirical evaluation of the presented method instruction enhancements consists of rooms to navigate between, other sets of enhancements can be used in different settings.

\subsection{Low-level Controller}
The controller $\pi_{c}(a_t|o_t,g_t,z_t)$ interacts with the environment through its primitive actions $a_t \in \mathcal{A}$. The controller expects on each timestep an egocentric RGB observation of the environment $o_t \in \Omega$ together with a task instruction $g_t \in \mathcal{G}$ and an instruction enhancement $z_t \in \mathcal{Z}$ provided by the meta-controller. The instruction informs the agent of its objective (e.g., \textit{find the red ball}), and the instruction enhancement (e.g., \textit{in the kitchen}) adds additional information on how the instruction should be carried out. The instruction enhancement will essentially navigate the agent to different rooms, resulting in episodic exploration of the different rooms in order to solve the main instruction. Both instruction and enhancement are provided using simple language sentences.

The action-space $\mathcal{A}$ of the controller consists of a discrete set of primitive movement steps (\textit{move forward, turn left, turn right}) and a special \textit{query}-action. This special query-action is invoked when the agent perceives itself near the goal object. Utilizing this action will invoke the goal assessment module.

Due to the utilization of instruction enhancements, the controller can be trained independent of the meta-controller. A straightforward way of training the controller, is to enhance the instructions by utilizing an oracle. When this oracle provides the most useful enhancement (e.g., which room should the agent visit next to find the goal) the extrinsic reward signal can be utilized to reward the agent. For example in the setting of object navigation, controllers can be rewarded by utilizing the improvement in geodesic distance between the agent and the goal object.

\subsection{Goal Assessment Module}

To signal that the agent thinks it has completed the objective, it needs to use a special \textit{done}-action. Utilizing this action will typically end the episode. However, as we will empirically demonstrate in Section~\ref{sec:failure_modes}, incorrect usage of this action is one of the main failure modes appearing prior to the introduction of a goal assessment module. In contrast, when the \textit{done}-action does not terminate the episode, the agent trains considerable faster.

In order to integrate the goal assessment module, the done-action is removed from the action-space of the controller. Instead, a \textit{query}-action is added to this action space. This novel query action will not terminate the episode (soft termination), but will query the goal assessment module. If the goal assessment module deems that the instruction is satisfied, and the agent is close enough to the target object, the agent will utilize the original done-action.

Essentially, the controller is now able to focus on low-level navigation, and consult an expert (the goal assessment module) in order to handle language grounding of the instruction. 

In order to allow the agent to find objects it did not see during training, a novel goal assessment model can be trained independent of the controller and meta-controller. Which is useful, as training a controller and meta-controller is typically more computational expensive.

In order to collect training data for the goal assessment module a random policy can be used, collecting both examples with goal objects, and observations without any visible objects. For positive samples the correct positive class is utilized 50\% of the time, while in the remainder cases another random possible instruction is utilized, together with the negative class label. This allows balancing out positive and negative labels.

\section{Empirical Evaluation}

\subsection{Environment Description}

\begin{figure*}
\centering
\includegraphics[scale=0.2, height=5cm]{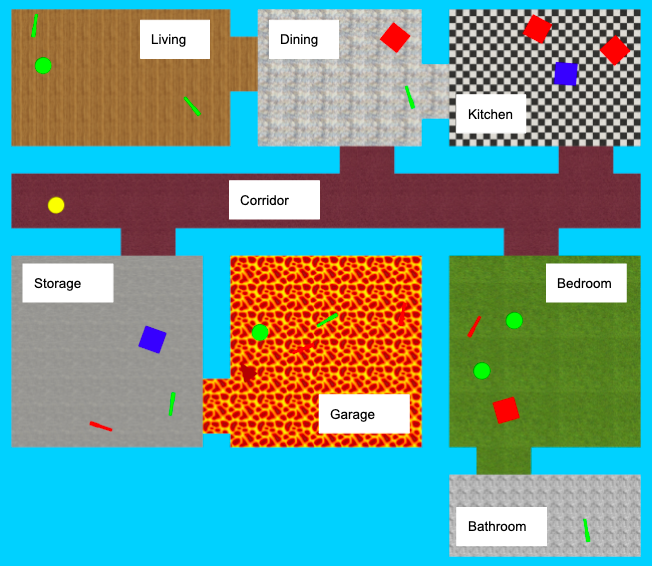}
\includegraphics[scale=0.2, height=5cm]{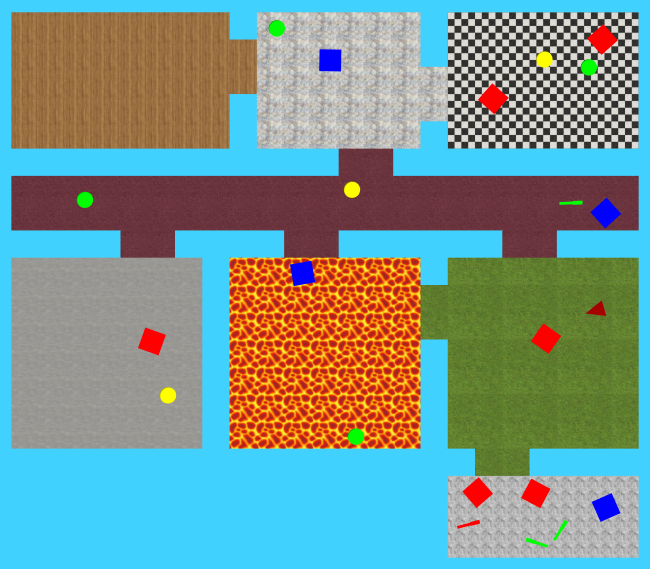}
\caption{Two different instances of the evaluation environment. Connections between rooms are randomized (with a holdout set of configurations). The agent has no access to this top-down map view.}
\label{fig:env}
\end{figure*}

In order to demonstrate the effectiveness of \gls{setie}, a simulated domestic environment is utilized within the \textit{MiniWorld} framework \cite{gym_miniworld}. Two instances are represented in Figure~\ref{fig:env}. The environment consists of 7 different rooms (garage, storage, bedroom, bathroom, living room, dining room and kitchen) together with a corridor that connects some of these rooms (depending on the instance). Each room has a distinctive look. As not all rooms are connected, the agent will often need to backtrack to previously visited points in order to further explore the environment.

Throughout the environment different abstract objects are randomly placed. Objects are defined by a category and a color. The categories used are \textit{box}, \textit{ball} and \textit{key}. In the experiments there is typically one goal object and multiple \textit{distractor} objects. In each task instance there is only a single object which matches the goal object description. The task is communicated using language through the template of \textit{"Find the [color] [shape]"}. The following objects are used during training: \textit{red box, green ball, blue box, yellow ball, red key and green key}. There is no association between objects and rooms.

On each timestep the agent observes an egocentric RGB observation $o_t$ of the environment. The reward function is densely defined, and consists of the improvement in the geodesic distance between the agent and the goal object. We use a slack penalty of $0.01$ which is subtracted from the reward on each timestep. When reaching the goal object we award the agent with a success bonus of 10.
\begin{equation}
    r_t(s_t,a_t, g_t) = (-\Delta_{geo\_dist} - 0.01) + 10*\mathbbm{1}_{success}
\end{equation}

Regarding actions, the agent is capable of turning left and right for a fixed amount, moving a fixed distance forward, and utilizing a special \textit{done}-action. In order to successfully complete an episode, the agent needs to use this done-action close to the goal object.

In each episode, the agent starts in a random position, and has no access to its current pose, the name of the room it is in, or a map of the environment. The connections between the different rooms are randomly enabled. However, each room is always accessible, and there are no uncommon connections (e.g., bathroom connected to kitchen). In total this results in 132 different possible floor plans. A holdout set of 30 floor plans is not utilized during training, but kept solely for evaluation purposes. This holdout set can be used in order to assess the generalization capabilities of the agent regarding floor plans.

\subsection{Baselines: why do non-hierarchical approaches fail?}
\label{sec:failure_modes}

\subsubsection{With soft-termination}

\begin{figure}[h!]
    \centering
    \includegraphics[scale=0.4]{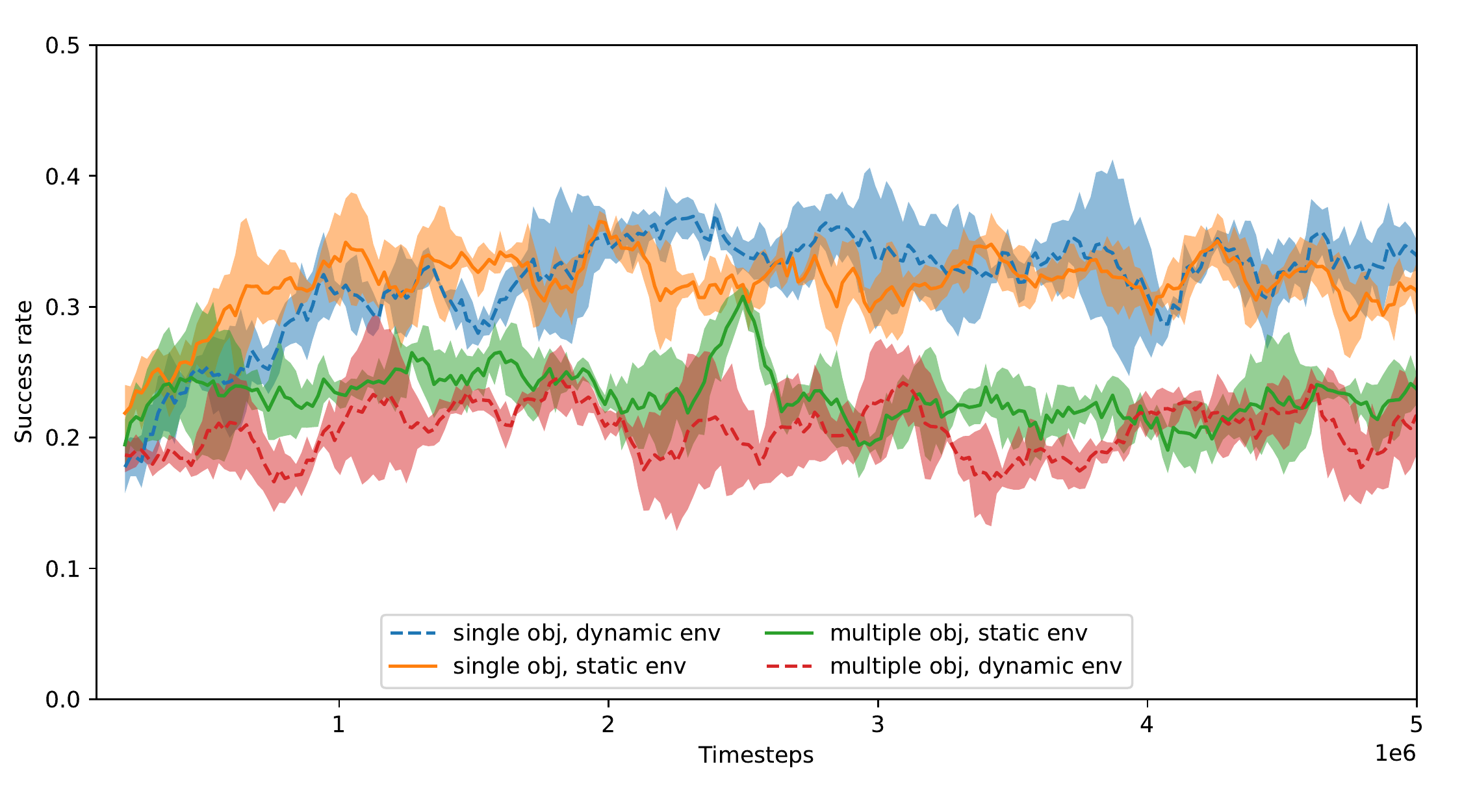}
    \caption{Training performance of a non-hierarchical PPO agent with soft-termination. Results are averaged over 3 runs.}
    \label{fig:baseline_training}
\end{figure}

When utilizing a non-hierarchical \gls{ppo} agent without any instruction enhancements, and with only a single object (a red box or blue box) the agent is capable of achieving an average success rate of $\sim35\%$ after 5 million interactions with the environment (Figure~\ref{fig:baseline_training}). When also introducing the problem of language grounding, by adding multiple objects to the environment, the agent has an average success rate of $\sim20\%$ after 5 million interactions.

\subsubsection{No soft-termination (full problem setting)}

If we also remove the relaxation of soft termination of the environment we arrive at the full problem setting. In this setting, when the agent utilizes the \textit{done}-action incorrectly, the episode is terminated. We analyzed the failure modes of the baseline agent in this setting (Figure~\ref{fig:failure_modes}):

\begin{figure}[h!]
    \centering
    \includegraphics[scale=0.5]{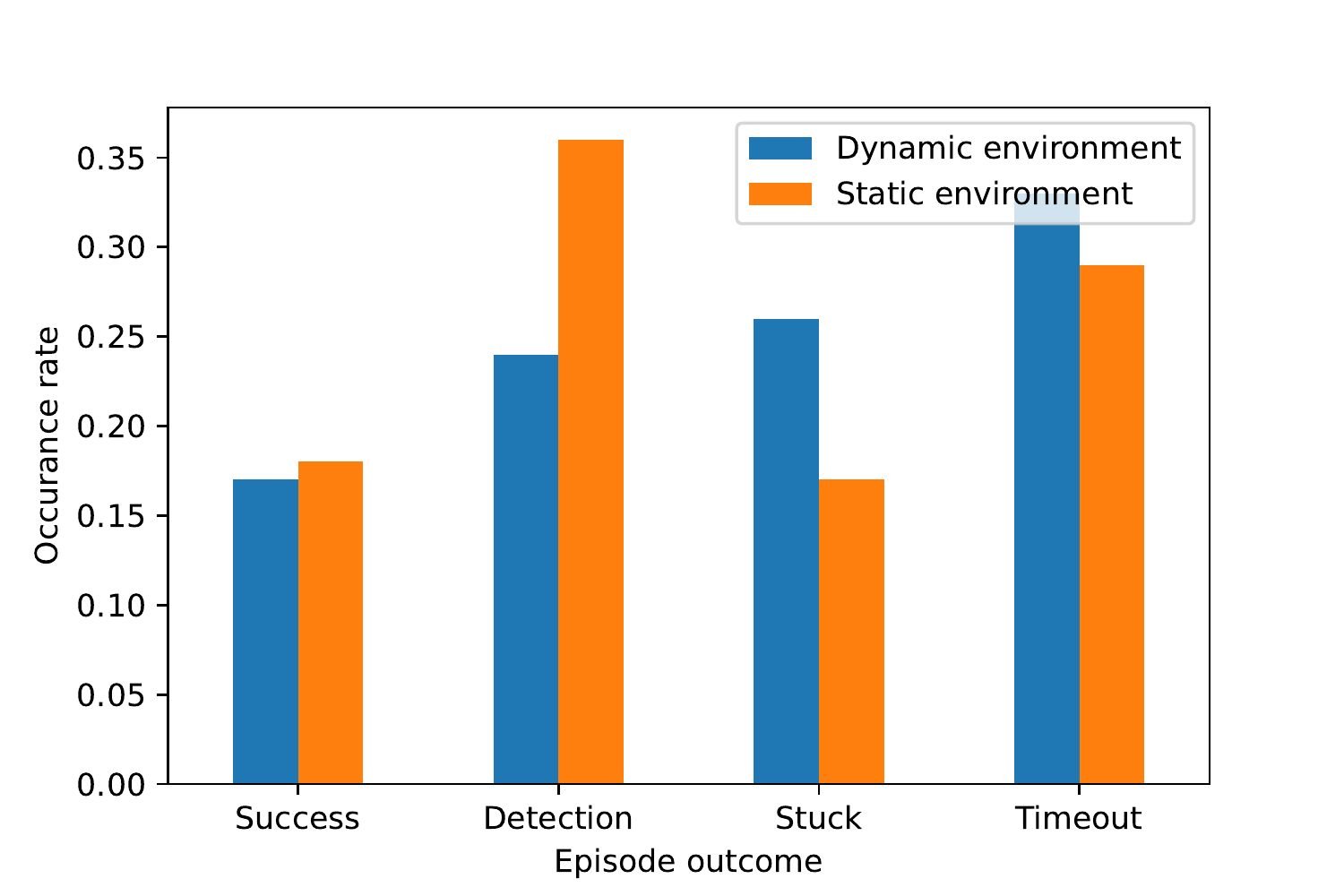}
    \caption{Failure modes of the trained non-hierarchical baseline. If the floor plan remains fixed (static environment), the amount of episodes where the agent gets stuck decreases, however this in turn increases goal detection errors.}
    \label{fig:failure_modes}
\end{figure}

\begin{itemize}
    \item \textbf{Detection:} agent used \textit{done}-action but was in the wrong position.
    \item \textbf{Timeout:} agent did not manage to find the goal within the allowed amount of timesteps, the agent did not use the \textit{done}-action at all.
    \item \textbf{Stuck:} distance between agent and goal object did not change in the final 10 steps.
\end{itemize}

When looking at these failure modes we noticed that the main reason for failure in a static environment setting, is related to the detection of goal objects. When also making the environment dynamic, both local navigation problems (getting stuck), and planning problems (timeout) start to occur more frequently.

\subsection{Does enhancing the instruction make the task more tractable?}

\begin{figure}[ht!]
    \centering
    \includegraphics[scale=0.4]{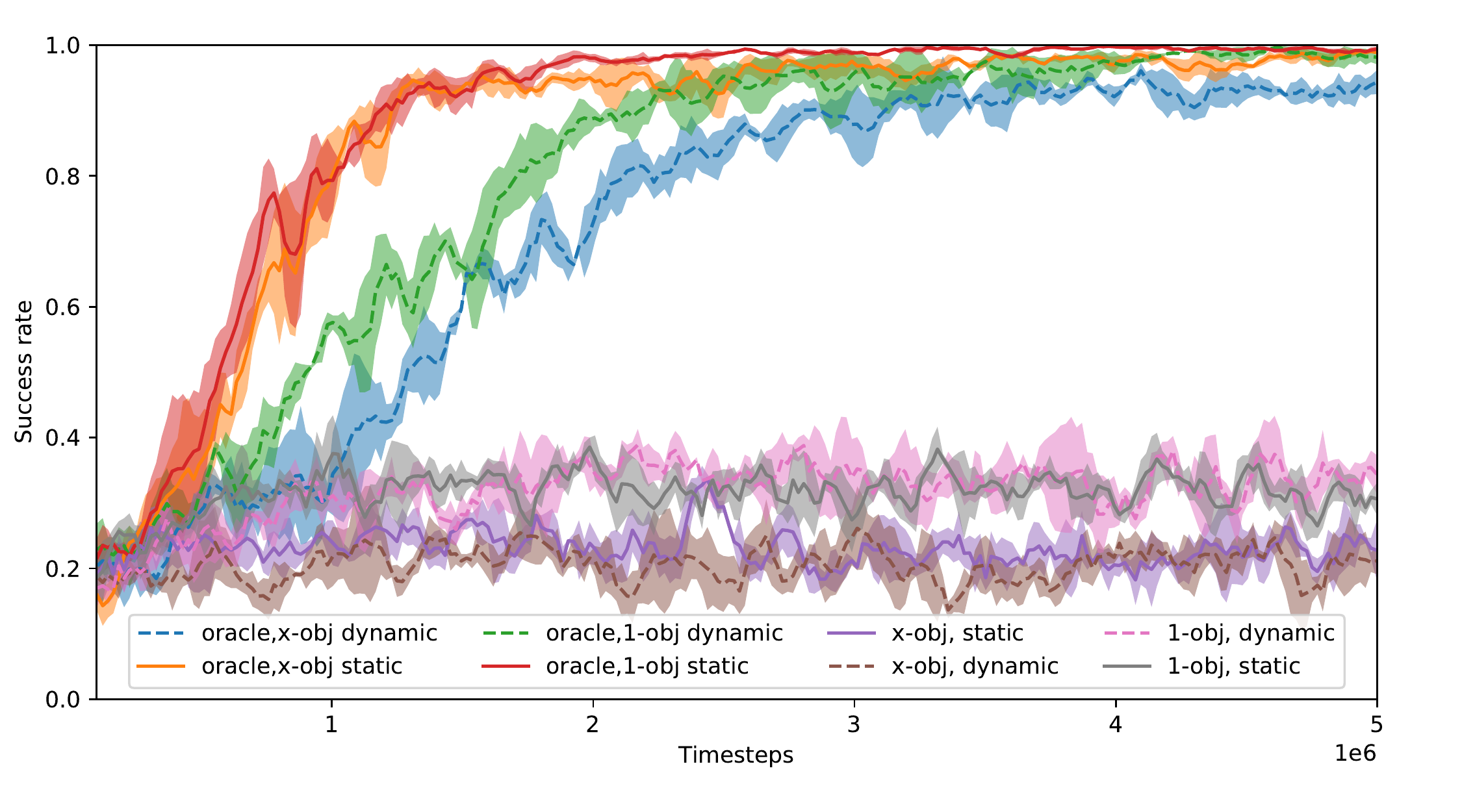}
    \caption{Training performance of the controller, in this setting the agent is allowed to use the done-action multiple times (soft termination). Without information to which room the agent should move next (oracle), the agent is unable to learn a policy in the environment. Results are averaged over 3 runs.}
    \label{fig:controller_training2}
\end{figure}

From the previous section, we can conclude that a non-hierarchical agent is not able to reliably solve the studied task. In order to validate whether enhancing the instruction will improve the performance, we trained an agent with its instructions enhanced through the use of an oracle. 

The utilized oracle is aware of the shortest path to the goal object in terms of rooms to visit. Having access to such an oracle outside the training environment, is an unrealistic assumption. The learned meta-controller will however take over the role of this oracle, providing adequate enhancements.

Utilizing an oracle based on the shortest path also alleviates the requirement of a custom reward function. If the controller is able to correctly interpret and follow the instruction enhancement, it will also collect the most reward.

As the results plotted in Figure~\ref{fig:controller_training2} indicate, enhancing the instructions allows the agent to almost entirely consistently solve the task both in the setting with a single object (1-obj) and multiple objects (x-obj). This validates the idea that enhancing the instruction allows the controller to carry out the low-level control task. In order to solve the entire task there is still the need to remove soft termination (Section~\ref{sec:soft_term}), and actually train a meta-controller (Section~\ref{sec:meta_train}).

\subsection{What is the impact of soft termination?}
\label{sec:soft_term}

\begin{figure}[ht!]
    \centering
    \includegraphics[scale=0.42]{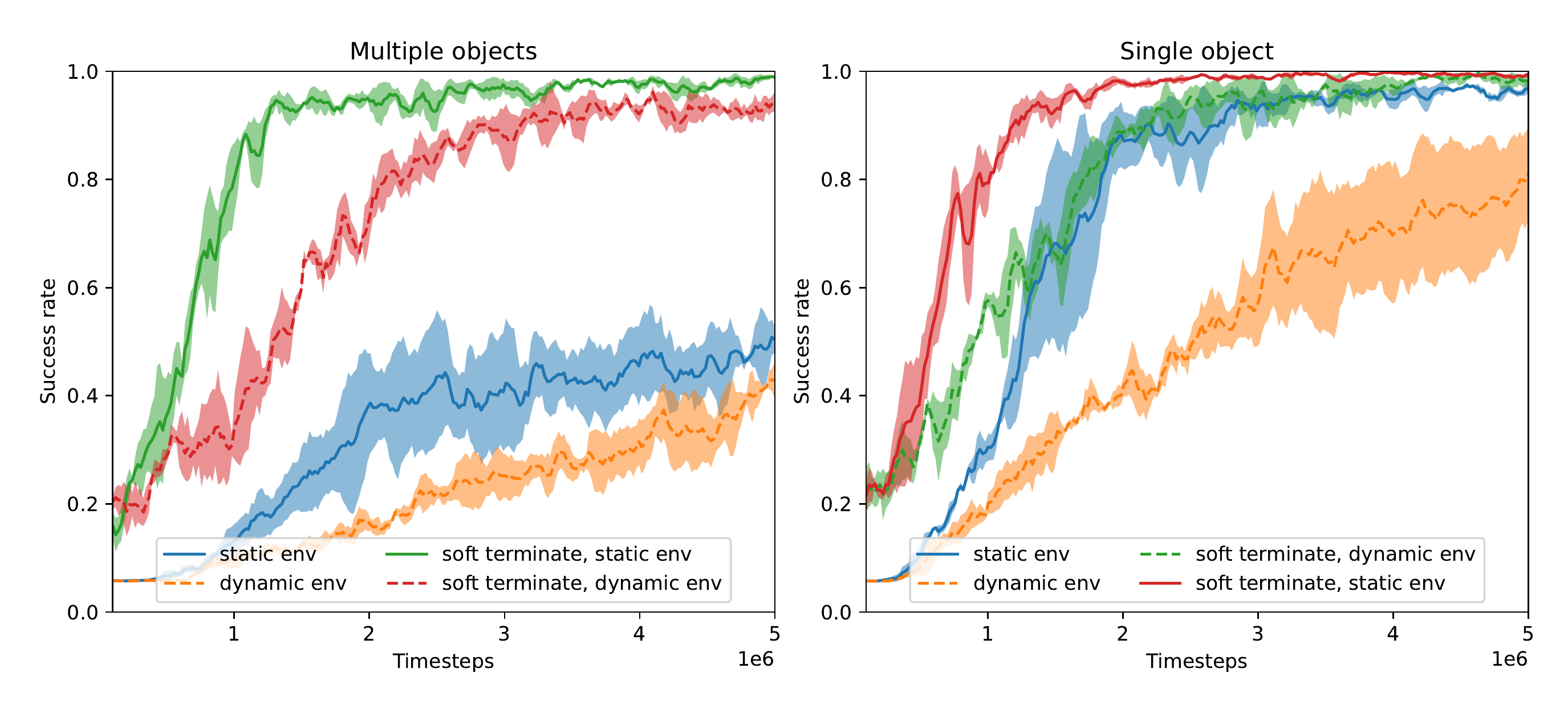}
    \caption{Training performance of the controller, with oracle instruction enhancements. Allowing soft termination, greatly improves sample efficiency. Results are averaged over 3 runs.}
    \label{fig:soft_terminate}
\end{figure}

In the previous experiments, the controller was trained using soft termination. This means that the agent is allowed to use the \textit{done}-action multiple times in an episode. Normally, this would terminate the episode, however we found that allowing the agent to utilize this action multiple times during training significantly increased the sample efficiency and success rate (Figure~\ref{fig:soft_terminate}). This training mechanism is especially crucial in the settings which require language grounding (multiple objects). We can allow this constraint due to the goal assessment module, which will filter out invalid done-actions when utilizing the entire architecture.

\subsection{Does a trained controller allow the meta-controller to solve the task?}
\label{sec:meta_train}

\begin{figure}[b!]
    \centering
    \includegraphics[scale=0.5]{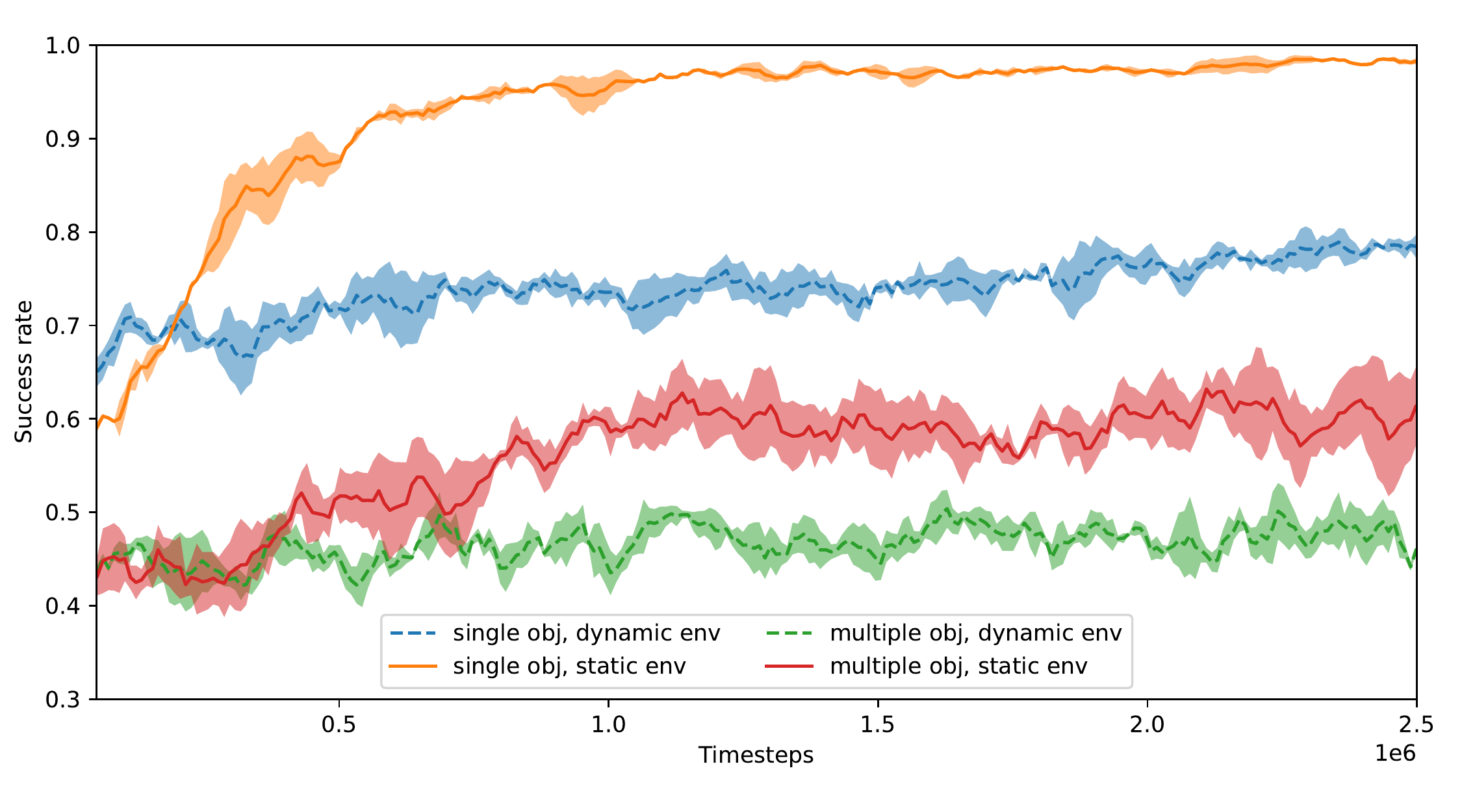}
    \caption{Meta-controller training performance. Results are averaged over 3 runs.}
    \label{fig:meta_controller_training}
\end{figure}

In Figure~\ref{fig:meta_controller_training} the results from training a meta-controller (through enhancing the instructions of a trained controller) in various configurations are plotted. The meta-controller has no problem exploring the environment when there is only a single static environment configuration used with a single goal object placed in it (SR $\sim95\%$). When multiple objects are present in the static environment setting, performance receives a significant hit (SR $\sim60\%$), but the agent is still able to improve its performance.

When the agent needs to manage dynamic instances of environments it starts with a high success rate, and is able to steadily improve (SR $\sim70\%$) in the setting with a single goal object. However, in the setting with both a dynamic environment configuration, and multiple objects the agent is not able to improve its initial performance (SR $\sim45\%$).

\subsection{Is the agent capable of exploring in a structured way?}

The failure modes of the baseline agent indicated that a lot of episodes ($\sim30\%$) failed due to the agent running out of allowed steps. This might indicate that the baseline agent is not able to explore the environment in a structured manner. In Table~\ref{table:exploration} we compare the percentage of the rooms the agent visited. From the results plotted in this table, we can conclude that the hierarchical approach is capable of covering a significantly larger proportion of the environment on average.

\begin{table}
    \centering
\begin{tabular}{|l|l|l|c|}
    \hline
    \textbf{Agent} & \textbf{Objects} & \textbf{Environment} & \textbf{Room coverage}      \\
    \hline
    Hierarchical & Single & Static &  51.0\% \\
         &      & Dynamic (holdout) &  45.4\% \\
         &      & Dynamic (train) &  45.5\% \\
         & Multiple & Static &  50.2\% \\
         &      & Dynamic (holdout) &  36.4\% \\
         &      & Dynamic (train) &  36.7\% \\
         \hline
    Flat (baseline) & Single & Static &  27.8\% \\
         &      & Dynamic (holdout) &  25.8\% \\
         &      & Dynamic (train) &  26.3\% \\
         & Multiple & Static &  12.5\% \\
         &      & Dynamic (holdout) &  12.6\% \\
         &      & Dynamic (train) &  12.6\% \\
    \hline
    \end{tabular}
    \caption{Average room coverage observed during evaluation runs.}
    \label{table:exploration}
\end{table}

\subsection{How well does the proposed hierarchical architecture performs?}

In this section the performance of the architecture is analyzed in its entirety. We are especially interested in how well the agent is capable of handling unseen environment floor plans, and novel objects.

\subsubsection{Zero-shot transfer to unseen environment configurations}

The agent is allowed to utilize 102 different floor plans during training. In order to validate whether the agent is capable of functioning in an environment it did not see during training, there is also a test-set containing 30 floor plans the agent did not see during training.

\begin{table}
    \centering
\begin{tabular}{ |l|c|c|c|c| } 
 \hline
 \textbf{Architecture} & \textbf{Objects} & \textbf{Static} & \textbf{Train} & \textbf{Test} \\ 
 \hline
 Flat PPO baseline & Single & $37\%\pm3.71$ & $42\%\pm4.79$ & $44\%\pm3.76$  \\ 
 Hierarchical + GA & Single & $81\%\pm5.20$ & $76\%\pm5.54$ & $75\%\pm4.67$  \\ 
 \hline
 Flat PPO baseline & Multiple &  $13\%\pm3.41$ &  $15\%\pm4.47$ & $12\%\pm2.66$ \\
 Hierarchical + soft term. & Multiple & $82\%\pm6.44$ & $69\%\pm5.76$ & $67\%\pm5.27$  \\ 
 Hierarchical & Multiple & $15\%\pm3.12$ & $18\%\pm2.81$ & $15\%\pm3.13$  \\ 
 Hierarchical + GA & Multiple & $52\%\pm3.06$ & $38\%\pm4.58$ & $40\%\pm4.52$  \\ 
 \hline
\end{tabular}
\caption{Overall performance of the entire architecture. For each setting 10 runs of each 100 random episodes where used.}
\label{table:overall_env_transfer}
\end{table}

From the results plotted in Table~\ref{table:overall_env_transfer} we can conclude that the hierarchical approach has a high success rate in the static environment. Especially, when there is no language grounding required.

In the setting with multiple objects, the hierarchical agent is now able to reach a high success rate when soft termination is allowed. When soft termination is disabled, the goal assessment module is capable of somewhat emulating this improved performance. However, there still remains room for improvement. When qualitatively looking at the mistakes made by the goal assessment module, we noticed that it often made mistakes if the goal object was barely visible in the single passed RGB observation.

In all cases, the agent was successfully capable of achieving a similar level of performance in the floor plan holdout set as in the training set.

\subsubsection{Zero-shot transfer to  unseen goal objects}

Because the instructions are formulated in natural language, we have an interface that makes it straightforward to test how well the agent handles combinations of colors and objects it did not see during training. The goal assessment module was retrained in order to be capable to detect the novel combinations of colors and shapes, while keeping all original navigation policies (controller and meta-controller).

\begin{table}
    \centering
\begin{tabular}{ |l|l|l|l| } 
 \hline
 \textbf{Environment:} & \textbf{Static} & \textbf{Train} & \textbf{Test} \\ 
 \hline
 Flat PPO Baseline & $15\%\pm1.69$ & $15\%\pm3.1$ & $14\%\pm4.21$  \\ 
 Hierarchical & $17\%\pm2.54$ & $15\%\pm2.96$ & $14\%\pm3.52$  \\ 
 Hierarchical + soft term. & $78\%\pm3.75$ & $69\%\pm2.75$ & $66\%\pm3.77$  \\ 
 Hierarchical + GA & $52\%\pm4.36$ & $39\%\pm3.7$ & $38\%\pm4.58$  \\
 \hline
\end{tabular}
\caption{Overall performance of the entire architecture on a holdout set of goal objects. For each setting 10 runs of each 100 random episodes where used.}
\label{table:overall_goal_transfer}
\end{table}

Similar to the zero-shot environment transfer experiments, we empirically can validate from the results in Table~\ref{table:overall_goal_transfer} that the agent is able to successfully find combinations of colors and shapes the agent did not see before without having to re-train the controller and meta-controller.

\vfill

\section{Related work}

\textbf{Object Navigation in RL} \\
Prior proposed architectures either fully rely on end-to-end training \cite{wijmans2020DDPPO}, make use of self-supervised learning through auxiliar tasks \cite{jaderberg2017aux_tasks,ye2020aux_tasks}, or use a planning-style approach by inferring maps from observations \cite{chaplot2020sem_exploration,gupta2020cognitive_mapping}. In contrast to our method, prior work relies on a pose sensor. \\

\noindent \textbf{Structured exploration through HRL} \\
HRL \cite{make4010009} is a core mechanism in object navigation. In these architectures a top-level meta-controller is trained to output relative goal position points which should be reachable by a trained PointGoal agent. For example the agent in \cite{narasimhan2020roomnav_maps} is capable of inferring a rough floor plan of the environment, the top-level outputs pointgoals in order to reach a desired area. This is similar to how we navigate the agent to different rooms in order to solve ObjectNav tasks.

Instead of using points as the interface between different levels of the architecture, natural language has also been proposed as the interface \cite{jiang2019hal,huHierarchicalDecisionMaking2019}, allowing the lower-level to generalize to unseen instructions. \\

\noindent \textbf{Language grounding} \\
The problem of language grounding has been approached solely from data \cite{misraMappingInstructionsVisual2017}, by adding auxiliary tasks and curriculum learning \cite{hermannGroundedLanguageLearning2017} and feature-wise affine transformation based on the instruction \cite{chevalierboisvert2019babyai}.

\section{Conclusion}
In this paper we study the problem of structured exploration in an object navigation setting. We demonstrate how the three sub-problems of: navigation, high level reasoning, and language grounding each contribute to the complexity of object navigation. A hierarchical approach is proposed in order to handle both the low-level navigation, and high-level planning. In order to have a loose coupling between the layers, language is used to enhance the original instruction in a way that makes it feasible for a low-level controller to partially tackle the overall task. To handle the third sub-problem of basic language grounding, a goal assessment module is introduced in order to guide the controller in assessing whether goal objects have been reached.

The effectiveness of the proposed architecture is empirically demonstrated in a simulated domestic environment. We demonstrate that the agent is able to better handle unseen environment configurations, and unseen goal objects compared to a non-hierarchical baseline.

In future work we plan on researching how we can further improve the performance in dynamic environments, make the set of instruction enhancements more dynamic, and how well the SETIE approach performs in real-world settings.

\section*{ACKNOWLEDGMENT}
This research received funding from the Flemish Government under the “Onderzoeksprogramma Artificile Intelligentie (AI) Vlaanderen” programme.

\bibliographystyle{splncs04}
\bibliography{references}

\begin{thebibliography}{10}
\providecommand{\url}[1]{\texttt{#1}}
\providecommand{\urlprefix}{URL }
\providecommand{\doi}[1]{https://doi.org/#1}

\bibitem{chaplot2020sem_exploration}
Chaplot, D.S., Gandhi, D., Gupta, A., Salakhutdinov, R.: Object {{Goal
  Navigation}} using {{Goal}}-{{Oriented Semantic Exploration}}. In: Advances
  in {{Neural Information Processing Systems}} 33 (2020)

\bibitem{chenAskYourHumans2021}
Chen, V., Gupta, A., Marino, K.: Ask {{Your Humans}}: {{Using Human
  Instructions}} to {{Improve Generalization}} in {{Reinforcement Learning}}.
  In: {{ICLR21}} (2021)

\bibitem{gym_miniworld}
Chevalier-Boisvert, M.: gym-miniworld environment for openai gym.
  \url{https://github.com/maximecb/gym-miniworld} (2018)

\bibitem{chevalierboisvert2019babyai}
{Chevalier-Boisvert}, M., Bahdanau, D., Lahlou, S., Willems, L., Saharia, C.,
  Nguyen, T.H., Bengio, Y.: {{BabyAI}}: First {{Steps Towards Grounded Language
  Learning With}} a {{Human In}} the {{Loop}}. In: {{ICLR19}} (2019)

\bibitem{cho2014gru}
Cho, K., van Merrienboer, B., Gulcehre, C., Bahdanau, D., Bougares, F.,
  Schwenk, H., Bengio, Y.: Learning {Phrase} {Representations} using {RNN}
  {Encoder}-{Decoder} for {Statistical} {Machine} {Translation}. In: {EMNLP14}
  (2014)

\bibitem{gupta2020cognitive_mapping}
Gupta, S., Tolani, V., Davidson, J., Levine, S., Sukthankar, R., Malik, J.:
  Cognitive {{Mapping}} and {{Planning}} for {{Visual Navigation}}.
  International Journal of Computer Vision  \textbf{128}(5),  1311--1330 (2020)

\bibitem{hermannGroundedLanguageLearning2017}
Hermann, K.M., Hill, F., Green, S., Wang, F., Faulkner, R., Soyer, H.,
  Szepesvari, D., Czarnecki, W.M., Jaderberg, M., Teplyashin, D., Wainwright,
  M., Apps, C., Hassabis, D., Blunsom, P.: Grounded {{Language Learning}} in a
  {{Simulated 3D World}}. arXiv:1706.06551 [cs, stat]  (2017)

\bibitem{huHierarchicalDecisionMaking2019}
Hu, H., Yarats, D., Gong, Q., Tian, Y., Lewis, M.: Hierarchical {{Decision
  Making}} by {{Generating}} and {{Following Natural Language Instructions}}.
  In: {{NeurIPS19}} (2019)

\bibitem{make4010009}
Hutsebaut-Buysse, M., Mets, K., Latré, S.: Hierarchical reinforcement
  learning: A survey and open research challenges. Machine Learning and
  Knowledge Extraction  \textbf{4}(1),  172--221 (2022)

\bibitem{jaderberg2017aux_tasks}
Jaderberg, M., Mnih, V., Czarnecki, W.M., Schaul, T., Leibo, J.Z., Silver, D.,
  Kavukcuoglu, K.: Reinforcement {{Learning}} with {{Unsupervised Auxiliary
  Tasks}}. In: {{ICLR17}} (2017)

\bibitem{jiang2019hal}
Jiang, Y., Gu, S., Murphy, K., Finn, C.: Language as an {{Abstraction}} for
  {{Hierarchical Deep Reinforcement Learning}}. In: {{NeurIPS19}} (2019)

\bibitem{karkus2021diff_slam}
Karkus, P., Cai, S., Hsu, D.: Differentiable slam-net: Learning particle slam
  for visual navigation. In: Proceedings of the IEEE/CVF Conference on Computer
  Vision and Pattern Recognition. pp. 2815--2825 (2021)

\bibitem{mishkin2019nav_benchmark}
Mishkin, D., Dosovitskiy, A., Koltun, V.: Benchmarking {Classic} and {Learned}
  {Navigation} in {Complex} {3D} {Environments}. arXiv:1901.10915 [cs]  (2019),
  \url{http://arxiv.org/abs/1901.10915}

\bibitem{misraMappingInstructionsVisual2017}
Misra, D., Langford, J., Artzi, Y.: Mapping {Instructions} and {Visual}
  {Observations} to {Actions} with {Reinforcement} {Learning}. In: Proceedings
  of the {Conference} on {Empirical} {Methods} in {Natural} {Language}
  {Processing} (2017)

\bibitem{nachum2019hrl_works}
Nachum, O., Tang, H., Lu, X., Gu, S., Lee, H., Levine, S.: Why {Does}
  {Hierarchy} ({Sometimes}) {Work} {So} {Well} in {Reinforcement} {Learning}?
  In: {NeurIPS} 2019 {DeepRL} {Workshop} (2019),
  \url{http://arxiv.org/abs/1909.10618}

\bibitem{narasimhan2020roomnav_maps}
Narasimhan, M., Wijmans, E., Chen, X., Darrell, T., Batra, D., Parikh, D.,
  Singh, A.: Seeing the {{Un}}-{{Scene}}: Learning {{Amodal Semantic Maps}} for
  {{Room Navigation}}. In: {{ECCV20}} (2020)

\bibitem{savva2019habitat}
Savva, M., Kadian, A., Maksymets, O., Zhao, Y., Wijmans, E., Jain, B., Straub,
  J., Liu, J., Koltun, V., Malik, J., Parikh, D., Batra, D.: Habitat: {A}
  {Platform} for {Embodied} {AI} {Research}. In: Proceedings of the {IEEE}
  {International} {Conference} on {Computer} {Vision} (2019)

\bibitem{schulman2017ppo}
Schulman, J., Wolski, F., Dhariwal, P., Radford, A., Klimov, O.: Proximal
  {Policy} {Optimization} {Algorithms}. arXiv:1707.06347 [cs]  (2017),
  \url{http://arxiv.org/abs/1707.06347}

\bibitem{szot2021habitat2}
Szot, A., Clegg, A., Undersander, E., Wijmans, E., Zhao, Y., Turner, J.,
  Maestre, N., Mukadam, M., Chaplot, D., Maksymets, O., Gokaslan, A., Vondrus,
  V., Dharur, S., Meier, F., Galuba, W., Chang, A., Kira, Z., Koltun, V.,
  Malik, J., Savva, M., Batra, D.: Habitat 2.0: {Training} {Home} {Assistants}
  to {Rearrange} their {Habitat}. In: Advances in {Neural} {Information}
  {Processing} {Systems}. vol.~34, pp. 251--266 (2021)

\bibitem{weihs2020allenact}
Weihs, L., Salvador, J., Kotar, K., Jain, U., Zeng, K.H., Mottaghi, R.,
  Kembhavi, A.: {AllenAct}: {A} {Framework} for {Embodied} {AI} {Research}. In:
  {CoRR2020} (2020)

\bibitem{wijmans2020DDPPO}
Wijmans, E., Kadian, A., Morcos, A., Lee, S., Essa, I., Parikh, D., Savva, M.,
  Batra, D.: {{DD}}-{{PPO}}: Learning {{Near}}-{{Perfect PointGoal Navigators}}
  from 2.5 {{Billion Frames}}. In: {{ICLR20}} (2020)

\bibitem{ye2020aux_tasks}
Ye, J., Batra, D., Wijmans, E., Das, A.: Auxiliary {Tasks} {Speed} {Up}
  {Learning} {PointGoal} {Navigation}. In: Proceedings of the 2020 {Conference}
  on {Robot} {Learning} (2020)

\end{thebibliography}

\end{document}